\documentclass{article}

\PassOptionsToPackage{numbers, compress}{natbib}
\usepackage[preprint]{neurips_2024}

\usepackage[utf8]{inputenc} % allow utf-8 input
\usepackage[T1]{fontenc}    % use 8-bit T1 fonts
\usepackage{url}  % simple URL typesetting
\usepackage{booktabs}  % professional-quality tables
\usepackage{amsfonts}  % blackboard math symbols
\usepackage{nicefrac}  % compact symbols for 1/2, etc.
\usepackage{microtype} % microtypography
\usepackage[colorlinks]{hyperref}  % hyperlinks
\hypersetup{
  colorlinks,
  citecolor=blue,
  linkcolor=red,
  urlcolor=blue}

\usepackage{graphicx}
\usepackage[table,xcdraw]{xcolor}
\usepackage{booktabs}
\usepackage{multirow}
\usepackage{makecell}

\usepackage{amsmath} 
\usepackage{enumitem}

\usepackage{algorithm}
\usepackage{algorithmic}

\usepackage{pifont}
\newcommand{\cmark}{\ding{51}}
\newcommand{\xmark}{\ding{55}}

\usepackage{wrapfig}
\usepackage{caption}
\usepackage{subcaption}

\usepackage[table,xcdraw]{xcolor}
\graphicspath{{figures/}}

\title{Chain of Agents: Large Language Models Collaborating on Long-Context Tasks}

\author{%
  Yusen Zhang$^{\clubsuit}$\thanks{Work done while the author was a student researcher at Google Cloud AI Research.} , 
  Ruoxi Sun$^\diamondsuit$,  
  Yanfei Chen$^\diamondsuit$,  
  Tomas Pfister$^\diamondsuit$,  
  Rui Zhang$^{\clubsuit\dagger}$,  
  Sercan Ö. Arik$^\diamondsuit$\thanks{ Last authors}  \\
  $^\clubsuit$ Penn State University, $^\diamondsuit$ Google Cloud AI Research\\
  \texttt{\{yfz5488, rmz5227\}@psu.edu, \{ruoxis, yanfeichen, 
 tpfister, soarik\}@google.com} \\
}

\begin{document}

\maketitle

\begin{abstract}
Addressing the challenge of effectively processing long contexts has become a critical issue for Large Language Models (LLMs). Two common strategies have emerged: 1) reducing the input length, such as retrieving relevant chunks by Retrieval-Augmented Generation (RAG), and 2) expanding the context window limit of LLMs. However, both strategies have drawbacks: input reduction has no guarantee of covering the part with needed information, while window extension struggles with focusing on the pertinent information for solving the task. To mitigate these limitations, we propose \textit{Chain-of-Agents (CoA)}, a novel framework that harnesses multi-agent collaboration through natural language to enable information aggregation and context reasoning across various LLMs over long-context tasks. CoA consists of multiple worker agents who sequentially communicate to handle different segmented portions of the text, followed by a manager agent who synthesizes these contributions into a coherent final output. CoA processes the entire input by interleaving reading and reasoning, and it mitigates long context focus issues by assigning each agent a short context. We perform comprehensive evaluation of CoA on a wide range of long-context tasks in question answering, summarization, and code completion, demonstrating significant improvements by up to 10\% over strong baselines of RAG, Full-Context, and multi-agent LLMs.
\end{abstract}

\section{Introduction}

Despite their impressive performance across a wide range of scenarios, LLMs struggle with tasks that involve long contexts~\citep{brown2020language,srivastava2022beyond,reid2024gemini}. Numerous application scenarios demand extremely long contexts, such as question answering~\citep{yang-etal-2018-hotpotqa, ho-etal-2020-constructing, trivedi-etal-2022-musique}, document and dialogue summarization~\citep{huang-etal-2021-efficient,zhong-etal-2021-qmsum,zhang2021exploratory,zhang2021summ,chen-etal-2022-summscreen}, and code completion~\citep{guo2023longcoder,liu2023repobench}, where the inputs contain entire books~\citep{10.1162/tacl_a_00023,kryscinski2021booksum} and long articles~\citep{dasigi-etal-2021-dataset}.

To tackle the challenge with long context tasks, two major directions have been explored as shown in Table~\ref{tab:compare}: {\it input reduction} and {\it window extension}.
{\it{Input reduction}} reduces the length of the input context before feeding to downstream LLMs. Truncation approaches~\citep{achiam2023gpt,team2023gemini} directly truncate the input. Retrieval Augmented Generation (RAG)~\citep{xu2023retrieval} extends this direction by retrieving the most relevant chunks through embedding similarity. However, because of low retrieval accuracy,  LLMs could receive an incomplete context for solving the task, hurting performance.
{\it Window extension} extends the context window of LLMs via finetuning to consume the whole input~\citep{chen2023extending,ma2024megalodon,munkhdalai2024leave}. For example, Claude-3~\citep{anthropic2024claude} directly allows reading 200k tokens for each input.
However, when the window becomes longer, LLMs struggle to focus on the needed information to solve the task, suffering from ineffective context utilization such as the {\it ``lost in the middle''} issue~\citep{li2024long,an2024make,liu2024lost}. 

\begin{table}[H]
\caption{Comparison between Chain-of-Agents and prior methods for long-context tasks. Rec./Foc.: being able to mitigate inaccurate receptive field/long context focusing issues.
Read: the number of tokens as model input, where $n$ is the total input length, $k$ is the context window limit of LLMs. Inter.: the interpretability of the approach. Note that RAG is `medium interpretable' because of the re-ranked chunks.}
\label{tab:compare}
\resizebox{\linewidth}{!}{
\begin{tabular}{@{}llcccccccl@{}}
\toprule
Category & Example Work & Rec. & Foc. & No Train & Read & Agent & Applicability & Inter. \\ \midrule
\multirow{2}{*}{Input Reduction}  & Truncation~\citep{OpenAI_GPT4_2023}   & \xmark  & \cmark    & \cmark & $k$   & Single   & Generic & Low    \\
    & RAG~\citep{xu2023retrieval}& \xmark  & \cmark    & \xmark  & $n+k$   & Single   & Query-based & Medium \\ \midrule
\multirow{2}{*}{Window Extension} & Position Interpolation~\citep{chen2023extending} & \cmark & \xmark& \xmark  & $n$    & Single   & Generic & Low    \\
    & Long Context~\citep{anthropic2024claude} & \cmark & \xmark& \xmark  & $n$   & Single   & Generic & Low    \\ \midrule
Multi-agent LLMs   & Chain-of-Agents (Ours)& \cmark & \cmark    & \cmark & $n$& Multiple    & Generic & High   \\ \bottomrule
\end{tabular}}
\end{table}

\begin{figure}[t!]
\centering
\includegraphics[width=\linewidth]{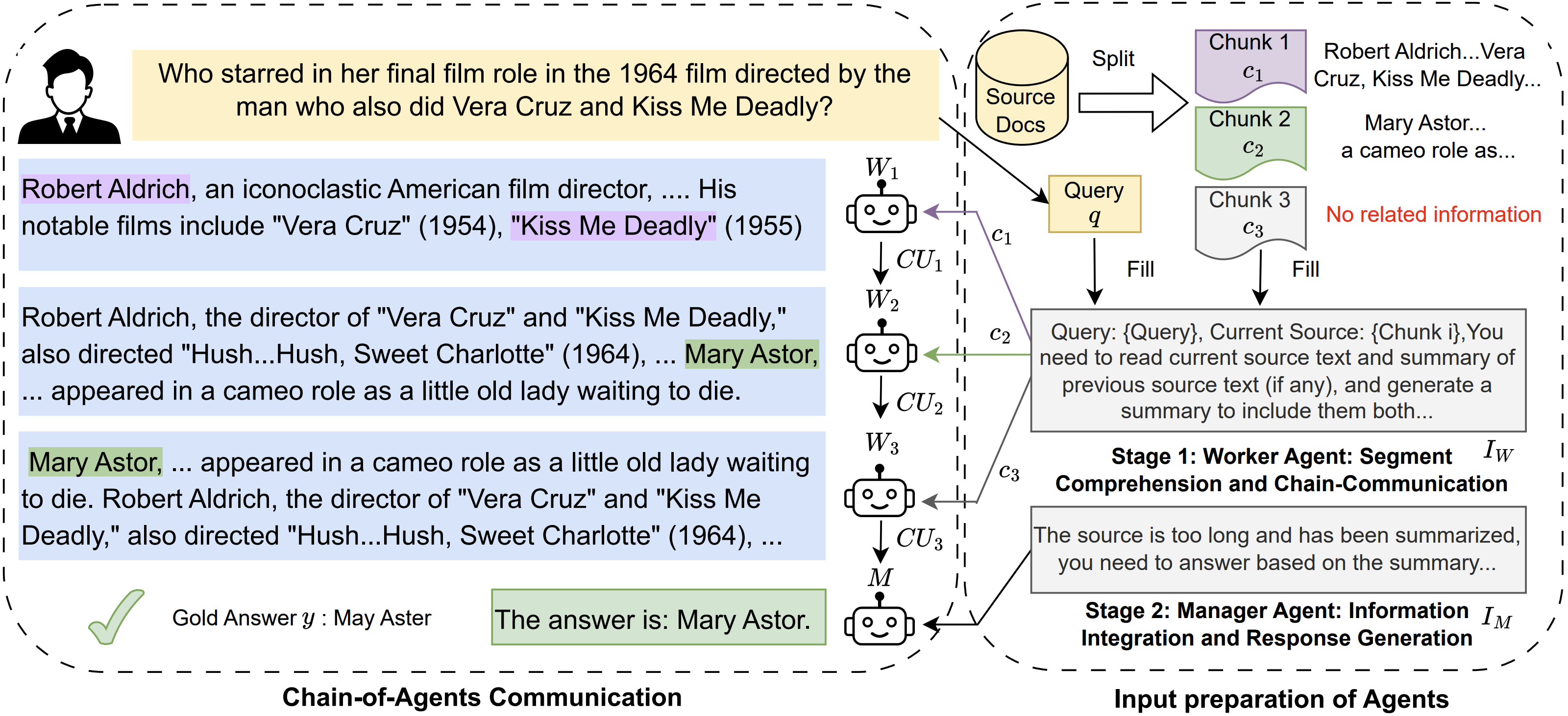}
\vspace{-5mm}
\caption{Overview of Chain-of-Agents, a training free, task agnostic, and highly-interpretable framework that harnesses multi-agent collaboration for long-context tasks. It consists of multiple worker agents who sequentially communicate to handle different segmented portions of the text, followed by a manager agent who synthesizes these contributions into a coherent final output.}
\vspace{-3mm}
\label{fig:coa}
\end{figure}

Motivated by the aforementioned challenges, we propose a novel framework, {\it Chain-of-Agents (CoA)}, inspired by human-like processing of long-context tasks. The key idea of CoA is to harness multi-agent communication to enable information aggregation and context reasoning capabilities across different LLMs. 
As shown in Figure~\ref{fig:coa}, CoA contains two stages. 
In stage 1, a series of worker agents in charge of different chunks of long context collaborate and aggregate evidence for answering the given query. To this end, the workers read and process sequentially, each receiving the message from previous worker and transferring the useful updated information to the next. In stage 2, the manager agent receives the complete evidence from last worker agent and generates the final response.

As shown in Table~\ref{tab:compare}, CoA is a training free, task agnostic, and highly interpretable framework processing entire ``receptive field'' by interleaved reading-processing and mitigating the long context focusing issue by assigning each agent a short context. Different from {\it input reduction} where LLMs need to start processing with low receptive field over reduced inputs  (``read-then-process''), workers in CoA start to process each chunk before reading all input (``interleaved read-process''), tackling the problems that input reduction struggles with, such as, generic summarization or counting of passages~\cite{bai2023longbench}. Different from {\it context extension}, CoA leverages the capability of communication rather than trying to feed many tokens into an LLM. This is a more natural solution for complex tasks because we assume that each LLM has its limit and there are always complex context tasks surpassing its limit.
Compared with Full-Context, CoA is also cost effective by reducing time complexity from $n^2$ to $nk$, where $n$ is input tokens and $k$ is the context limit of LLMs.

We conduct intensive experiments on {\it nine datasets}, including question answering, summarization, and code completion tasks with {\it six LLMs}, with PaLM 2~\citep{anil2023palm}, Gemini~\citep{team2023gemini}, and Claude 3~\citep{anthropic2024claude} models. We compare CoA with two strong baselines chosen from \textit{input reduction} and \textit{window extension} approaches, respectively: (i) RAG, which uses a state-of-the-art retriever to obtain the most relevant information to feed into the LLM and (ii) Full-Context (Vanilla), which feeds all input into the LLM until reaching the window limit. Our results show that on all nine datasets, CoA obtains significant improvement over all baselines by up to 10\%. Noting that there is not enough research on multi-agent for long context tasks, we carefully create two multi-agent baselines, including a hierarchical structure and result merging approach to further demonstrate that CoA is superior among other possible multi-agent frameworks.

\section{Related work}

\paragraph{Multi-agent LLMs.} Multi-agent LLMs has become a popular topic~\citep{guo2024large}. A large proportion of works focus on social simulation. ``Generative agents'' is a sandbox environment allowing 25 agents to communicate with each other~\citep{park2023generative}, while ``Social simulacra'' extends to 1000 agents~\citep{park2022social}. Based on the success of them, some works explore the game settings~\citep{light2023avalonbench,wang2023avalon,xu2023exploring,xu2023language,mukobi2023welfare}, world wars~\citep{hua2023war}, economy markets~\citep{li2023large,weiss2023rethinking}, recommendation systems~\citep{zhang2023generative}, and pandemics~\citep{ghaffarzadegan2023generative}. Others advance problem solving, focusing on reasoning of short text via multi-agent debating~\citep{du2023improving,xiong2023examining,chen2023multi,tang2023medagents} and discussing~\citep{chen2023reconcile,saha2023branch} for different tasks in reasoning~\citep{du2023improving,tang2023medagents}, mechanics problems~\citep{ni2024mechagents}, paper review~\citep{xu2023towards}, knowledge graph construction~\citep{ye2023beyond}, and code intelligence~\citep{wang2023mac,huang2023agentcoder}. Different from the above works, we improve problem solving on long context tasks. To the best of our knowledge, the closest work utilizes a tree structure to do single-hop QA over long context~\citep{chen2023walking}. However, it is not designed for multi-hop reasoning or other tasks without communication between sibling agents. 

\paragraph{Long Context Modeling for LLMs.} \textit{Input Reduction}: RAG is broadly leveraged to solve long context query-based tasks~\citep{xu2023retrieval,ai2023information}. Combined with a strong retriever~\citep{bge_embedding,lin-etal-2023-train,wang2022text}, LLMs are expected to handle long context questions in open domains. Previous studies have augmented LLMs during pretraining~\citep{izacard2023atlas, wang-etal-2023-shall}, finetuning~\citep{lewis2020retrieval}, inference~\citep{yogatama-etal-2021-adaptive}, or directly integrating~\citep{jiang-etal-2022-retrieval,shi2023replug}. Moreover, some token-level retrieval approaches are proposed~\citep{li2023unlocking}. Longllmlingua~\citep{jiang2023longllmlingua} removes tokens from long prompt to compress long context prompt to a desired budget.  \textit{Window Extension}: The context windows of LLMs are getting longer and longer thanks to the development of GPUs. For instance, the context window increases from 1024 (GPT-2~\citep{Radford2019LanguageMA}), 2048 (GPT-3~\citep{NEURIPS2020_1457c0d6}), to 128k (GPT-4~\citep{OpenAI_GPT4_2023}). Moreover, the newest version of Claude-3~\citep{anthropic2024claude} supports 200k context windows. To save the cost of LLM training, some continue learning or finetuning approaches are proposed to extend the context window of pretrained LLMs~\citep{mohtashami2023landmark,press2021train,ma2024megalodon, munkhdalai2024leave}. For instance, position interpolation~\citep{chen2023extending} modifies rotary position encoding~\citep{su2024roformer} and extends the context length of LLaMA~\citep{touvron2023llama} to 32k. Different from the above works, CoA does not reduce the input length or extend the window length of LLMs, but rather leverages multi-agent collaboration and communication to obtain the full receptive field.

\paragraph{Complex Task Reasoning.} Previous works on complex reasoning have focused on decomposing the complex question into sub-questions to solve them step-by-step. \cite{perez-etal-2020-unsupervised} decompose the questions with an unsupervised model and answer them separately with another model. Decomposed Prompting~\citep{Khot2022DecomposedPA} leverages some predefined modules to classify each decomposed sub-question, then further decompose if needed. Additionally, decomposing is used for human-computer interaction~\citep{10.1145/3491102.3517582}, and prompter training~\citep{wang-etal-2022-iteratively}. Recently, many work has been proposed for LLMs, such as Chain-of-thought~\citep{Wei2022ChainOT} Least-to-most prompting~\citep{Zhou2022LeasttoMostPE} and Pearl~\citep{Sun2023PEARLPL}. However, the length of the prompt does not exceed the context limit of a single agent. By contrast, our Chain of Agents framework is proposed to effectively reason across multiple agents to support the unlimited length of source text.
\begin{wrapfigure}{R}{0.5\textwidth}
\begin{minipage}{0.5\textwidth}
\vspace{-8mm}
\begin{algorithm}[H]
\small
	\caption{Chain of Agents (CoA).}
	\label{alg:coa}
	\begin{algorithmic}
		\REQUIRE Source input $x$, query $q$, agent window size $k$, \\
		         \quad \quad \; large language model $\text{LLM}(*)$.
		\ENSURE Answer to the query.
		\STATE Split $x$ into $l$ chunks $\{c_1,c_2,\cdots,c_l\}$
		\STATE where $c_i$ is shorter than $k$
		\STATE Initialize $CU_0$ $\gets \text{empty string} $.
   \FOR {$i$ in $1,2,\cdots,l$}
  \STATE $CU_{i} \gets \text{LLM}_{W_i}(I_W, CU_{i-1}, c_i, q)$
   \ENDFOR    
		\RETURN $\text{LLM}_M(I_M,CU_{l},q)$
	\end{algorithmic}
\end{algorithm}
\end{minipage}
\end{wrapfigure}

\section{Method}

Figure~\ref{fig:coa} shows the overview of our Chain-of-Agents (CoA) framework, containing two stages. In stage 1, long context is split into chunks where each chunk can be processed by a worker agent. Then, the worker agents communicate sequentially to produce evidence over the entire context. In stage 2, a manager agent consumes the knowledge from the chain of workers to generate the final answer.

To formulate the task, we denote a long-context sample as $(x,y,q)$, where $x$ is the input of $n$ tokens, $y$ is the output of $m$ tokens, $q$ is an optional query. 
Given a LLM with $k$ tokens (usually $k \ll n$) as the context window limit, the target is to generate $y$ with the limited input context window. Therefore, we divide each source text $x$ into chunks $ x= \{c_1, c_2 ...c_l\}$, so that each chunk can be completely fed into the LLM agent backbone model.

\subsection{Stage 1: Worker Agent: Segment Comprehension and Chain-Communication}
 
In Stage 1, CoA contains a sequence of $l$ number of worker agents. Each worker $W_i$ inputs the concatenation of a chunk $c_i$ from source text $x$, a query $q$, instruction for a specific task for worker agent $I_W$, and the message passed from the previous agent, denoted as ``communication unit'' $CU_{i-1}$. The worker agents process them and output the message $CU_{i}$ for next worker, expressed as:
\begin{equation}
    CU_{i} = \text{LLM}_{W_i}(I_W, CU_{i-1}, c_i, q), 
\end{equation}

CUs produced by worker agents vary across different tasks. For question answering, CU contains the evidence for the manager to answer the question. For summarization, CU contains the summary of the previous texts. For code completion, CU contains the code summary with function/class names and explanation. Effectiveness on diverse tasks demonstrates the flexibility of CoA (Appendix~\ref{sec:case}).

The multi-step worker communication in CoA expands the model context to the full receptive field, meaning that the last worker can read the full input no matter how long the input is. Therefore, CoA is extensible to inputs with different lengths by adjusting the number of worker agents.

The left side of Figure~\ref{fig:coa} underscores the necessity of collaborative communication among workers to effectively address complex, long-context reasoning tasks.
We observe that 1) Although the question is unanswerable given $c_1$, $W_1$ generates related evidence that is useful for answering the question; 2) with the partial answer from the previous worker, $W_2$ further reasons with the current source to complete the full reasoning chain across agents and generate the interpretative reasoning chain; 3) $W_3$ finds no related information in the chunk 3, it directly rewrites $CU_2$ by putting the correct answer as the first token of $CU_3$ without adding any unrelated information. 
This shows that if workers are independent (such as tree structure communication), it is impossible to answer hop two while the answer of hop one is held by another worker. 

\subsection{Stage 2: Manager Agent: Information Integration and Response Generation}
In Stage 2, after multiple steps of information extraction and comprehension by worker agents, the manager agent produces the final solution. While worker agents extract relevant information in a long-context source, the manager agent synthesizes relevant information accumulated by the end of {\it ``worker-agent-chain''} to generate the final answer. 
Specifically, given the instruction for manager $I_M$ and query $q$, the manager agent consumes accumulated knowledge from last worker $CU_{l}$ and generates the final answer $Response$:
\begin{equation}
    Response = \text{LLM}_M(I_M,CU_{l},q)
\end{equation}
The benefit of using a separate LLM as the manager agent is to decompose the duty of analyzing chunks in the long-context source ({\it ``worker agents''}) and producing the final answer ({\it``manager agent''}), so that every agent can fulfill its duty to the most\footnote{Other design choices: Our experiments show that using the last worker $W_l$ to directly generate the final result leads to a performance drop. Besides, feeding the manager with all $CU_i$ or some $CU$ that is related to the answer (decided by $W_i$) also hurts the performance because of confusion led by conflicting $CU_i$.}.

\subsection{Time Complexity Analysis}
\begin{wraptable}{r}{5.3cm}
\vspace{-10mm}
    \centering
    \caption{Time complexity.}
    \begin{tabular}{@{}lll@{}}
    \toprule
   & Encode & Decode \\ \midrule
    Full-Context & $\mathcal{O}(n^2)$ & $\mathcal{O}(nr)$  \\
    CoA  & $\mathcal{O}(nk)$ & $\mathcal{O}(nr)$  \\ \bottomrule
\end{tabular}
\label{tab:complexity}
\end{wraptable}
We compare the time cost of full-context input and Chain-of-Agents theoretically in a decoder-only setting. We assume the response generated by LLMs contains $r$ tokens on average, the input has $n$ tokens, and the context limit of LLM is $k$. The time complexity is shown in Table~\ref{tab:complexity} (Appendix~\ref{sec:proof}). As can be seen, the encoding time of CoA is less than Full-Context because $k \ll n$ in long context tasks, while they have the same decoding time. This demonstrates the efficiency of CoA compared with the Full-Context baseline.
\vspace{-2mm}
\section{Experiment}
\vspace{-2mm}
\subsection{Experiment Setup}
\label{sec:setup}
\begin{table}[!t]
\caption{\textbf{Dataset Statistics}. Avg. Input/Agents is the average words/agents (8k) for source input.}
\resizebox{\linewidth}{!}{
\begin{tabular}{@{}lccccccccc@{}}
\toprule
\multicolumn{1}{l}{} & \multicolumn{5}{c}{Question Answering} & \multicolumn{3}{c}{Summarization} & \multicolumn{1}{c}{Code} \\ \cmidrule(lr){2-6}\cmidrule(lr){7-9}\cmidrule(lr){10-10}
     & HotpotQA & MuSiQue  & NarrativeQA & Qasper  & QuALITY & QMSum    & GovReport & BookSum & RepoBench-P    \\ \midrule
Avg. Input    & 10603 & 12975 & 71787    & 4236 & 4936 & 12524 & 9239    & 108478 & 7105\\
Avg. Agents& 2.35     & 2.88     & 12.45 & 1.12    & 1.31    & 2.57     & 2.03   & 18.63 & 1.69 \\
Query-based & \cmark & \cmark & \cmark & \cmark & \cmark & \cmark & \xmark  & \xmark & \cmark \\ \bottomrule
\end{tabular}}
\vspace{-4mm}
\label{tab:statis}
\end{table}
\paragraph{Datasets.} We conduct experiments on nine long context datasets across three task types (Table~\ref{tab:statis}):
\begin{itemize}[leftmargin=*,noitemsep,topsep=0pt,parsep=0pt,partopsep=0pt]
    \item \textbf{Question Answering.} We consider five QA datasets from the LongBench~\citep{bai2023longbench} and SCROLL~\citep{shaham-etal-2022-scrolls}.
    \textbf{HotpotQA}~\citep{yang-etal-2018-hotpotqa} is a Wikipedia-based multi-hop QA dataset. It requires reasoning across multiple passages to find the answer. 
    \textbf{MuSiQue}~\citep{trivedi-etal-2022-musique} is a multi-hop QA dataset. It is much more difficult than HotpotQA as it contains more hops in one sample, unanswerable questions, and harder distracting content. \textbf{NarrativeQA}~\citep{kovcisky2018narrativeqa} is a QA dataset over entire books or movie transcripts. The answers can be abstract or extractive, yes/no, and unanswerable. 
    \textbf{Qasper}~\citep{dasigi2021dataset} is a question answering dataset over NLP papers. It also contains extractive, abstractive, yes/no, and unanswerable questions. 
    \textbf{QuALITY}~\citep{pang-etal-2022-quality} is a dataset based on stories and articles with multiple-choice questions for each sample. The model needs to select the correct answer among choices.
    \item \textbf{Summarization.} We pick two summarization datasets from SCROLLS. \textbf{QMSum}~\citep{zhong-etal-2021-qmsum} is a query-based summarization dataset, formed by meeting transcripts from multiple domains such as academic and industrial products. \textbf{GovReport}~\citep{huang-etal-2021-efficient} is a generic summarization dataset containing long reports published by the U.S. Government Accountability Office. We also use one dataset for long context memorization tasks. \textbf{BookSum}~\citep{kryscinski2021booksum} is a collection of datasets for long-form narrative summarization, including novels, plays, and stories. We use the book-level partition of the BookSum dataset for experiments.
    \item \textbf{Code Completion.} We pick one code completion dataset from LongBench. \textbf{RepoBench-P}~\citep{liu2023repobench} is collected from GitHub repositories, and the model needs to generate the next line of code given the long code base.
\end{itemize}
\paragraph{Metrics.} We report the geometric mean of ROUGE~\citep{lin-2004-rouge} for Summarization tasks, code similarity score~\citep{bai2023longbench} for Code Completion task, exact match for QuALITY dataset~\citep{shaham-etal-2022-scrolls}, and F1 score for the rest of the Question Answering datasets~\citep{bai2023longbench}.

\paragraph{LLMs.} We use six LLMs in total as the backbone of CoA across all experiments. \textbf{PaLM 2}~\citep{anil2023palm} is a series of models with a dense left-to-right, decoder-only language model pretrained on a high-quality corpus of 780 billion tokens. We use \textbf{text-bison@001} and \textbf{text-unicorn@001} for the experiments with an 8k maximum context window. \textbf{Gemini 1.0}~\citep{team2023gemini} is a family of LLMs proposed by Google. We use \textbf{gemini-ultra} for experiments. The input limit is 32k tokens for Gemini. \textbf{Claude 3}~\citep{anthropic2024claude} is a family of large language models developed by Anthropic. The family includes three state-of-the-art models in ascending order of capability: \textbf{claude-3-haiku}, \textbf{claude-3-sonnet}, and \textbf{claude-3-opus}. These models are capable of consuming 200k tokens in the context window, providing a strong baseline for long context tasks. Although our framework is flexible to use diverse types of LLMs as workers and manager, we use the same model for each $W_i$ and $M$ if not specified. 

\paragraph{Baselines.} Our principle of choosing baselines is to find the strongest and most typical approaches from \textit{input reduction} and \textit{window extension}. The first baseline is \textbf{Vanilla}. It directly consumes tokens until the context window of LLM is fully utilized, implying a 200k window LLM if using Claude 3. The other one is Retrieval-Augmented Generation (\textbf{RAG}). We use the state-of-the-art retriever~\citep{bge_embedding}. Following~\citep{xu2023retrieval}, we first segment the source text into chunks of 300 words and re-rank them using a retriever. Top-n chunks are then fed into the downstream LLM until the context window is fully utilized. GovReport dataset does not contain a query initially, we create a pseudo query ``What is the summary of the whole government report?'' as the query to rerank. 

To evaluate the performance of CoA compared with possible multi-agent approaches, we carefully construct two multi-agent approaches. For these two approaches, similar to CoA, we also assign each chunk $c_i$ to $W_i$ using similar instructions to generate $CU_i$. In these approaches, worker agents are parallel and independent while CoA is sequential. \textbf{Multi-Agent Voting} (\textbf{Merge}) Each agent directly generate an answer $a_i$ according to $c_i$. A majority voting is applied to all answers $a_i$ to decide the final answer. \textbf{Multi-Agent Hierarchical Structure} (\textbf{Hierarchical}). Inspired by~\citep{chen2023walking}, we propose a hierarchical framework, where the communication forms a tree structure between workers $W_i$ and manager $M$. For each worker, it first judges whether $c_i$ contains useful information. If true, it generates a communication unit $CU_i$. Then, all $CU_i$ are sent to the manager $M$ to come up with a final answer.  Besides, we append an integer number $L$ at the end of every approach to clearly remind the window size limit of LLM. For instance, ``CoA (8K)'' refers to the base LLM used in CoA with window size 8K.

\subsection{Overall Results of CoA}

\begin{table}[t!]
\caption{\textbf{Overall results of CoA}. CoA significantly outperforms Vanilla and RAG using various backbone LLMs on all datasets.}
\resizebox{\linewidth}{!}{
\begin{tabular}{@{}clcccccccc@{}}
\toprule
\multicolumn{1}{l}{}&   & \multicolumn{5}{c}{Question Answering}& \multicolumn{2}{c}{Summarization} & \multicolumn{1}{c}{Code} \\ \cmidrule(lr){3-7}\cmidrule(lr){8-9}\cmidrule(lr){10-10}
\multicolumn{1}{l}{LLMs} & \multicolumn{1}{l}{Baselines}  & \multicolumn{1}{l}{HotpotQA} & \multicolumn{1}{l}{MuSiQue} & \multicolumn{1}{l}{NarrativeQA} & \multicolumn{1}{l}{Qasper} & \multicolumn{1}{l}{QuALITY} & \multicolumn{1}{l}{QMSum} & \multicolumn{1}{l}{GovReport} & \multicolumn{1}{l}{RepoBench-P} \\\midrule
\multirow{3}{*}{text-bison}   & Vanilla (8k)  & 45.57  & 26.87 & 11.96 & 26.56& 61.86 & 15.45  & 20.60   & 56.30   \\
& RAG (8k)& 51.91  & 33.83 & 14.20 & 27.20& 55.28 & 15.59  & 20.83 & 55.63   \\
& CoA (8k)& \textbf{53.62} & \textbf{37.09}& \textbf{25.26}& \textbf{37.17} & \textbf{65.42}& \textbf{16.77}& \textbf{26.11}  & \textbf{58.25}  \\\midrule
\multirow{3}{*}{text-unicorn} & Vanilla (8k)  & 51.09  & 29.67 & 14.45 & 27.81& 83.40 & 16.61  & 23.50   & 53.87   \\
& RAG (8k)& 58.01  & 40.38 & 19.12 & 24.44& 83.00 & 16.83  & 21.43 & 50.49   \\
& CoA (8k)& \textbf{62.04} & \textbf{42.49}& \textbf{20.37}& \textbf{38.01} & \textbf{83.80}& \textbf{17.67}& \textbf{26.48}  & \textbf{60.39}  \\\midrule
\multirow{4}{*}{gemini-ultra} & Vanilla (8k)  & 40.62  & 23.61 & 7.71& 20.59& 57.40 & 12.10  & 26.18   & 49.09   \\
& Vanilla (32k) & 45.09  & 27.93 & 7.21& 21.71& 58.60 & 10.24  & 26.96   & 73.04   \\
& RAG (8k)& 51.13  & 31.56 & 14.51 & 18.70& 62.40 & 12.70  & 25.87& 72.94   \\
& CoA (8k)& \textbf{54.26} & \textbf{35.09}& \textbf{25.26}& \textbf{35.10} & \textbf{80.60}& \textbf{12.84}& \textbf{26.98}  & \textbf{73.05}  \\ \bottomrule
\end{tabular}}

\label{tab:main}
\end{table}

\paragraph{Question Answering.} Table~\ref{tab:main} shows the results of Question Answering tasks on all three models. CoA (8k) outperforms Vanilla (8k) by a large margin on \textit{all 8 datasets}, including 13.30\% on NarrativeQA, 12.82\% on MuSiQue, and 22.00\% on Quality, for text-bison, text-unicorn, and gemini-ultra, respectively. Also, CoA (8k) outperforms RAG (8k) model for all 8 datasets using all three LLMs, demonstrating that CoA achieves higher performance than RAG. In other words, \textbf{using multi-agent LLMs outperforms 
RAG models}. It is also worth noting that for gemini-ultra, Vanilla (32k) improves the Vanilla (8k) baseline, yet it is still lower than CoA (8k).
\paragraph{Summarization and Code Completion.} Table~\ref{tab:main} shows the results of Summarization and Code Completion tasks. Similarly, CoA (8k) also outperforms all Vanilla (8k) and (32k) baselines on all three datasets, demonstrating the strong capability of CoA on various tasks. It is worth noting that for GovReport, RAG fails to improve the baseline with pseudo query. By contrast, CoA improves the performance significantly, showing that \textbf{CoA can also be applied in non-query tasks}. 

\paragraph{Long Context LLMs.}
As Claude 3 models support 200k of tokens, we call these models long context models (LCM). Table~\ref{tab:long} shows the performance of the LCM on two datasets. As can be seen, CoA (8k) outperforms Vanilla (200k) significantly, showing that with only an 8k context window, \textbf{CoA can achieve a much higher performance than LCM with a 200k context window}. Also, CoA improves the performance with the samples that can be fed into a 200k context window (no truncation). Moreover, the improvements over the Vanilla (200k) and RAG (8k) become higher when the model size increases from Haiku to Opus (11.63/11.36/17.4 for NarrativeQA, 1.66/2.86/3.47 for BookSum). This demonstrates that \textbf{CoA benefits from stronger models to achieve higher improvements}. 

\begin{table}[t!]
\caption{Comparison with long context LLMs on NarrativeQA and BookSum. CoA significantly outperforms Claude 3 with 200k context limits. No Trun./Trun. indicates the source text in the sample is less/more than 200k tokens which does not need/needs truncation for vanilla (200k) baseline. Average is the mean value across all samples.}
\resizebox{\linewidth}{!}{
\begin{tabular}{@{}lccccccccc@{}}
\toprule
 & \multicolumn{3}{c}{claude-3-haiku}  & \multicolumn{3}{c}{claude-3-sonnet} & \multicolumn{3}{c}{claude-3-opus}\\\cmidrule(lr){2-4}\cmidrule(lr){5-7}\cmidrule(lr){8-10}
 & \multicolumn{1}{l}{No Trun.} & \multicolumn{1}{l}{Trun.} & \multicolumn{1}{l}{Average} & \multicolumn{1}{l}{No Trun.} & \multicolumn{1}{l}{Trun.} & \multicolumn{1}{l}{Average} & \multicolumn{1}{l}{No Trun.} & \multicolumn{1}{l}{Trun.} & \multicolumn{1}{l}{Average} \\\midrule
\textbf{NarrativeQA} & \multicolumn{1}{l}{} & \multicolumn{1}{l}{}    & \multicolumn{1}{l}{}   & \multicolumn{1}{l}{} & \multicolumn{1}{l}{}    & \multicolumn{1}{l}{}   & \multicolumn{1}{l}{}  & \multicolumn{1}{l}{}& \multicolumn{1}{l}{}   \\ 
Vanilla (200k)    & 8.00& 2.50   & 7.17  & 5.58& 2.44   & 5.15  & 7.23 & 2.35    & 6.56  \\
RAG (8k)& 5.94& 4.22   & 5.71  & 9.09& 5.17   & 8.50  & 6.13 & 4.29    & 5.86  \\
CoA (8k)& \textbf{18.31}    & \textbf{21.34}  & \textbf{18.80} & \textbf{16.63}   & \textbf{16.47} & \textbf{16.51} & \textbf{24.38}& \textbf{21.26}   & \textbf{23.96} \\\midrule
\textbf{BookSum}& \multicolumn{1}{l}{} & \multicolumn{1}{l}{}    & \multicolumn{1}{l}{}   & \multicolumn{1}{l}{} & \multicolumn{1}{l}{}    & \multicolumn{1}{l}{}   & \multicolumn{1}{l}{}  & \multicolumn{1}{l}{}& \multicolumn{1}{l}{}   \\
Vanilla (200k)    & 11.98    & 11.70  & 12.04 & 12.17    & 11.90  & 12.10 & 14.11& 13.67   & 14.00 \\
CoA (8k) & \textbf{13.28}   & \textbf{13.73}  & \textbf{13.70} & \textbf{14.92}    & \textbf{15.05}  & \textbf{14.96} & \textbf{17.74}& \textbf{16.68}   & \textbf{17.47} \\\bottomrule
\end{tabular}}
\vspace{-5mm}
\label{tab:long}
\end{table}

\paragraph{Other Multi-Agent Frameworks.}

As shown in Table~\ref{tab:other_baselines}, Hierarchical (8k) outperforms Vanilla (8k) on five out of eight datasets, demonstrating the hierarchical approach can also improve the vanilla baselines significantly. Merge (8k) is lower than Vanilla (8k) except for GovReport, showing that merging is especially effective for long summarization tasks such as GovReport. As can be seen, CoA outperforms Hierarchical and Merge on all eight datasets. The reason behind the results is because Hierarchical and Merge do not allow workers to communicate with each other due to their parallel designs. Thus, each worker can only maintain the information in its own chunk $c_i$ which blocks the understanding of the whole text, hurting the performance greatly. 

\begin{table}[t!]
\caption{Comparison between CoA and other multi-agent frameworks. CoA with sequential agents outperforms other designs with multiple parallel agents including Merge and Hierarchical.}
\label{tab:other_baselines}
\resizebox{\linewidth}{!}{
\begin{tabular}{@{}lcccccccc@{}}
\toprule
& \multicolumn{1}{l}{HotpotQA} & \multicolumn{1}{l}{MuSiQue} & \multicolumn{1}{l}{NarrativeQA} & \multicolumn{1}{l}{Qasper} & \multicolumn{1}{l}{QuALITY} & \multicolumn{1}{l}{QMSum} & \multicolumn{1}{l}{GovReport} & \multicolumn{1}{l}{RepoBench-P} \\ \midrule
Vanilla (8k)  & 45.57  & 26.87 & 11.96 & 26.56& 61.86 & 15.45  & 20.60   & 56.30  \\
Merge (8k)  & 42.96  & 26.66 & 11.27 & 26.78& 59.30 & 9.42& 25.38   & 33.66 \\
Hierarchical (8k) & 50.62  & 29.40 & 17.04 & 31.39& 64.20 & 15.19   & 16.54   & 27.96 \\
CoA (8k)& \textbf{53.62}  & \textbf{37.09} & \textbf{25.26} & \textbf{37.17}& \textbf{65.42} & \textbf{16.77}   & \textbf{26.11}   & \textbf{58.25} \\ \bottomrule
\end{tabular}}
\vspace{-5mm}
\end{table}
 \vspace{-5mm}
\section{Analyses}
 \vspace{-2mm}

\subsection{CoA Improvement is More Obvious When RAG Fails to Retrieve Gold Answer}

\begin{figure}[!t]
    \centering
    \begin{minipage}[t]{0.49\linewidth}
  \centering
  \includegraphics[width=\textwidth]{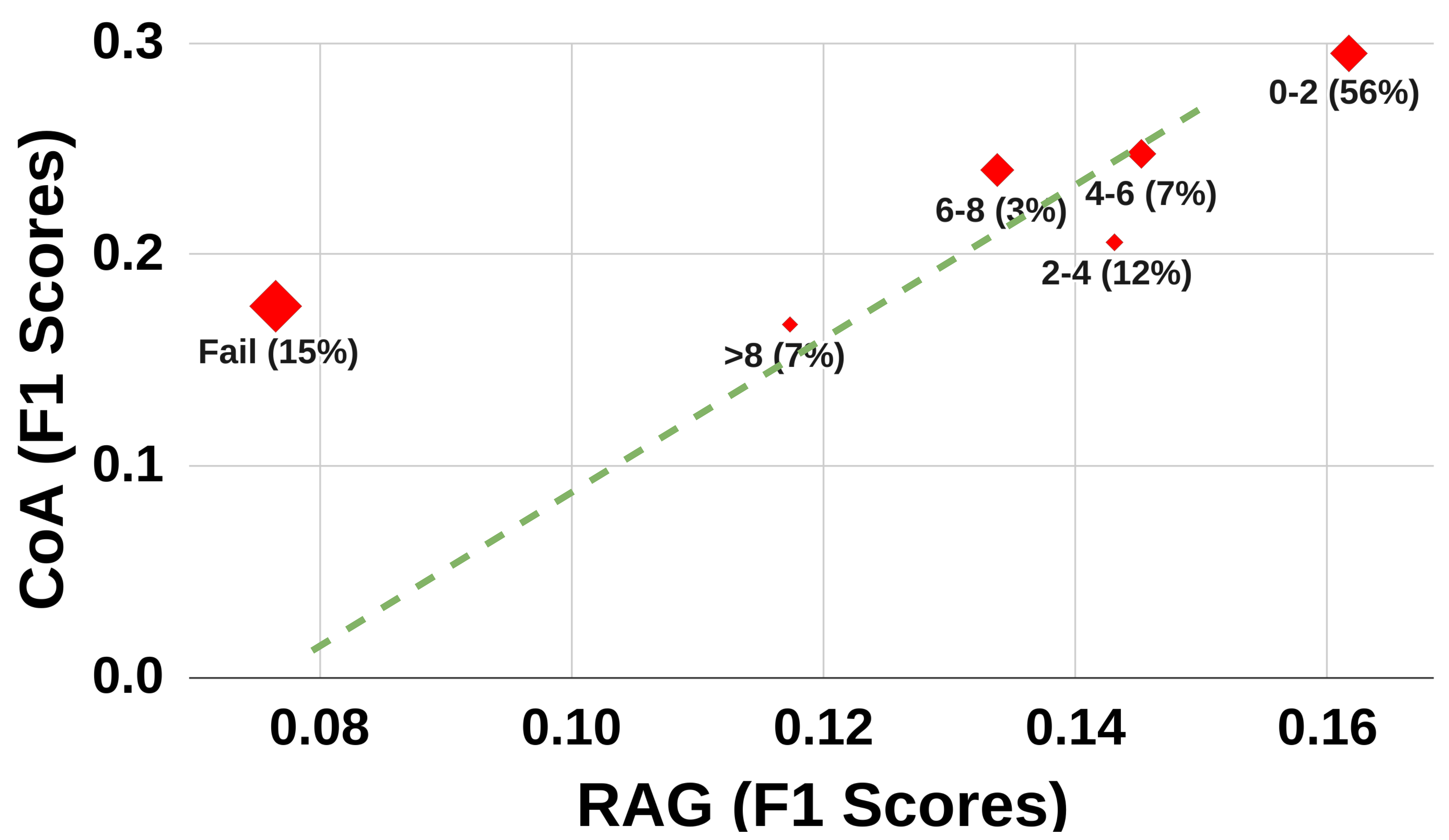}
  \vspace{-5mm}
  \caption{Comparison on NarrativeQA. X-axis/Y-axis indicate RAG/CoA performance while each point represents a bin. The number indicates the chunk index of gold answer (ratio of number of samples in bracket), and the size of the point indicates the improvement of CoA over RAG.  }
  \vspace{-2mm}
  \label{fig:coavsrag}
    \end{minipage}
    \hspace{1mm}
    \begin{minipage}[t]{0.49\linewidth}
  \centering
  \includegraphics[width=\linewidth]{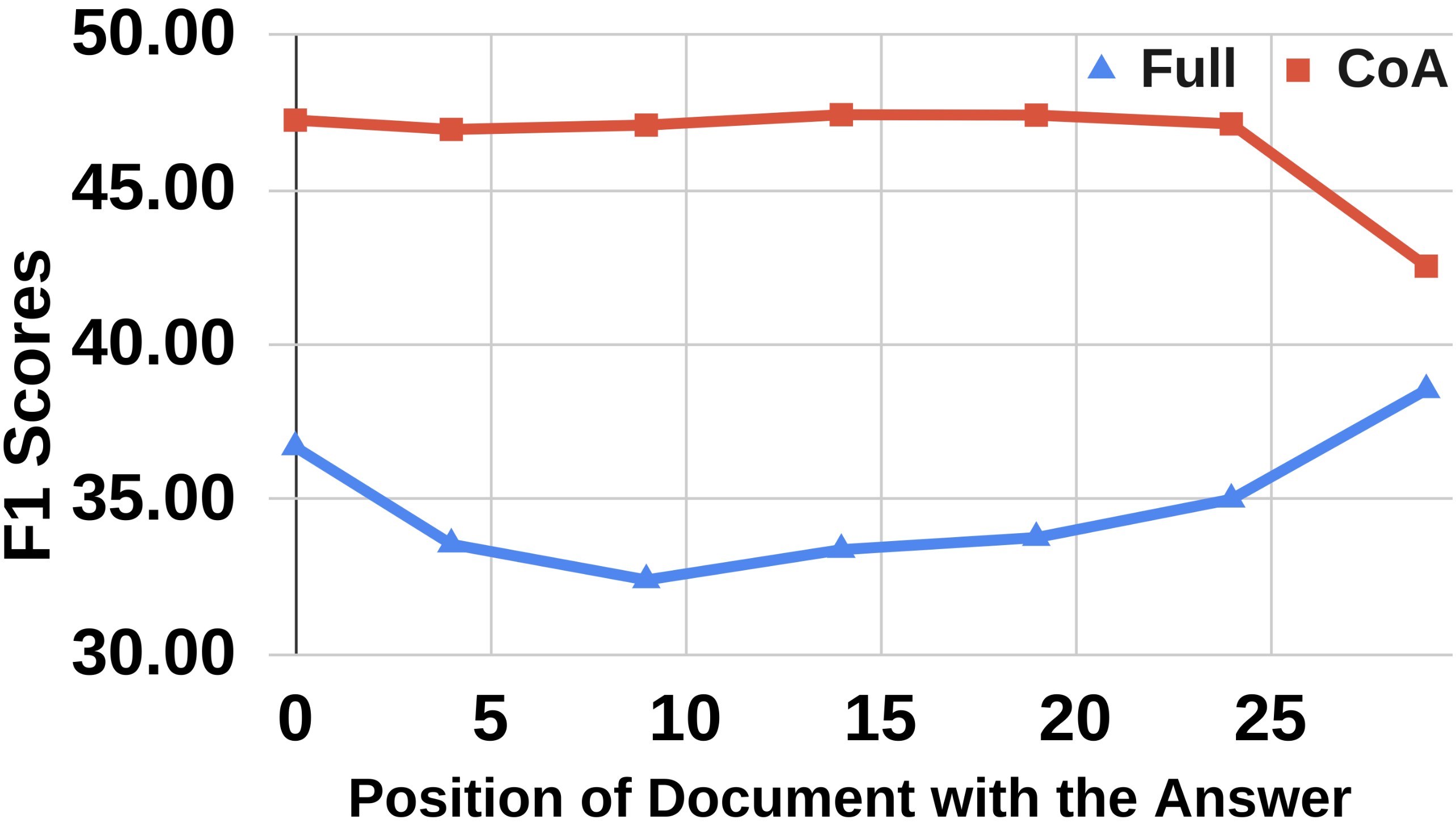}
  \vspace{-5mm}
  \caption{Performance of CoA and Full on Natural Questions. CoA mitigates the lost-in-the-middle issue. X-axis is the index of document with gold answer where small number indicates gold answer is closer to start.}
  \vspace{-2mm}
  \label{fig:lost}
    \end{minipage}
\end{figure}

To demonstrate this, we first classify the samples in NarrativeQA dataset into different bins according to the position (index) of the chunk in RAG processed input that contains the gold answer. Then, we compute the average score of the CoA and RAG results of different bins. Figure~\ref{fig:coavsrag} shows the results. As shown in the figure, RAG performs better when the index is smaller (the gold answer is nearer to the top), showing that downstream LLMs rely significantly on the quality of RAG re-ranking. Besides, the performance of RAG is positively correlated to CoA's when it successfully retrieves the gold answer. However, when RAG fails, CoA can greatly improve the performance (much higher than the tendency line).
\subsection{CoA Improvement is More Obvious When Long Context Models Meet Longer Inputs}

\begin{figure}
     \centering
     \begin{subfigure}[b]{0.32\textwidth}
         \centering
         \includegraphics[width=\textwidth]{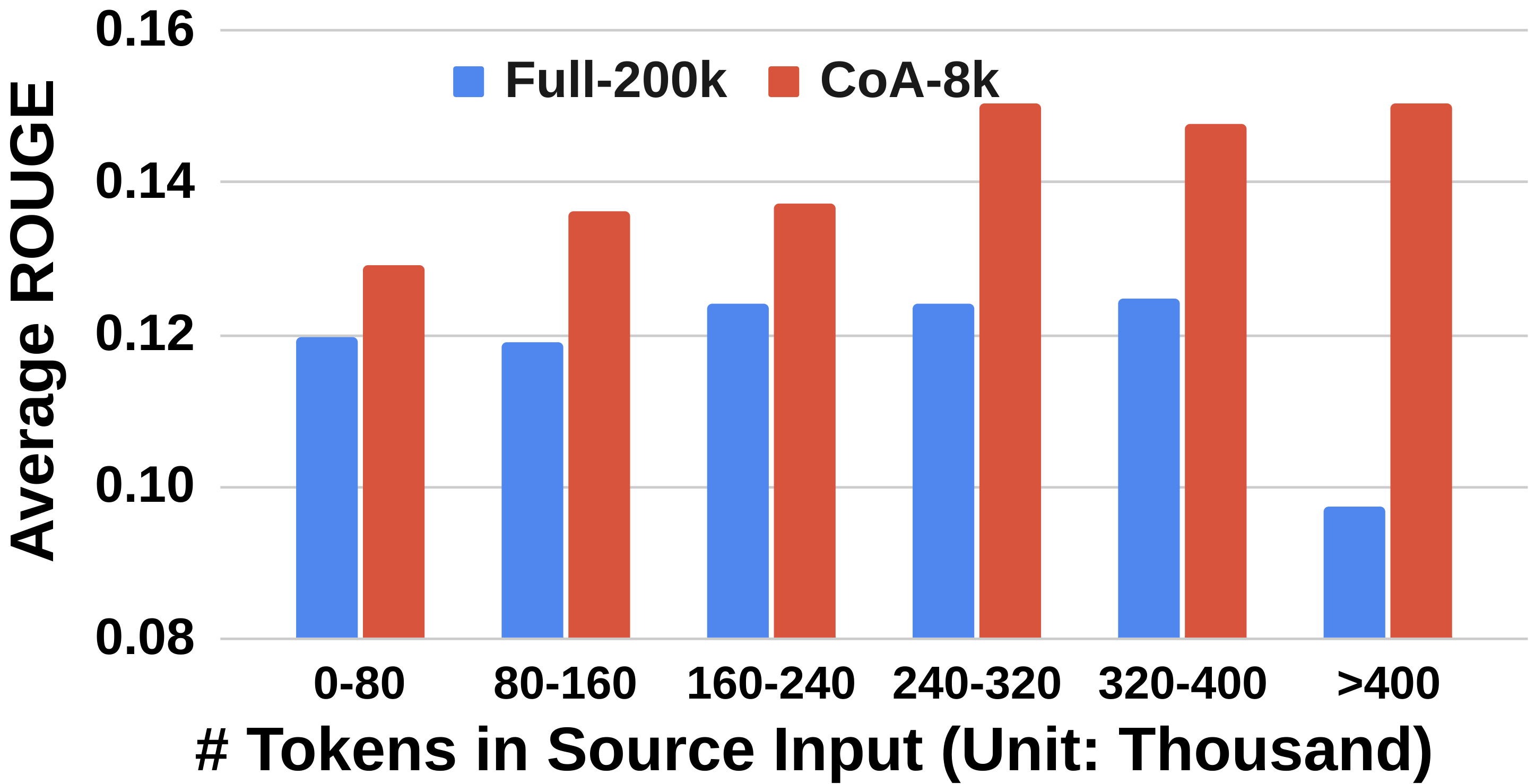}
         \caption{Claude 3 Haiku}
         \label{subfig:haiku}
     \end{subfigure}
     \hfill
     \begin{subfigure}[b]{0.32\textwidth}
         \centering
         \includegraphics[width=\textwidth]{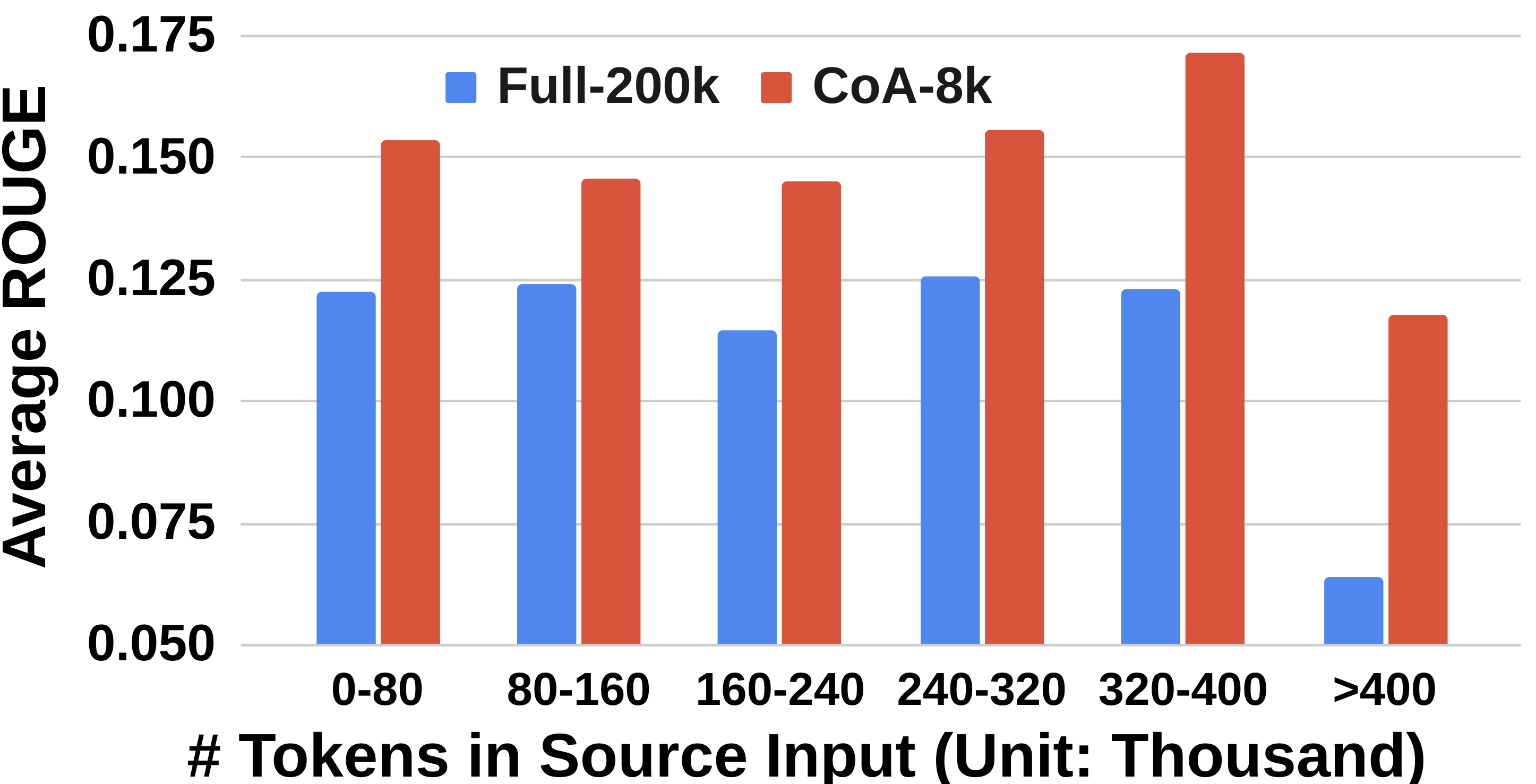}
         \caption{Claude 3 Sonnet}
         \label{subfig:sonnet}
     \end{subfigure}
     \hfill
     \begin{subfigure}[b]{0.32\textwidth}
         \centering
         \includegraphics[width=\textwidth]{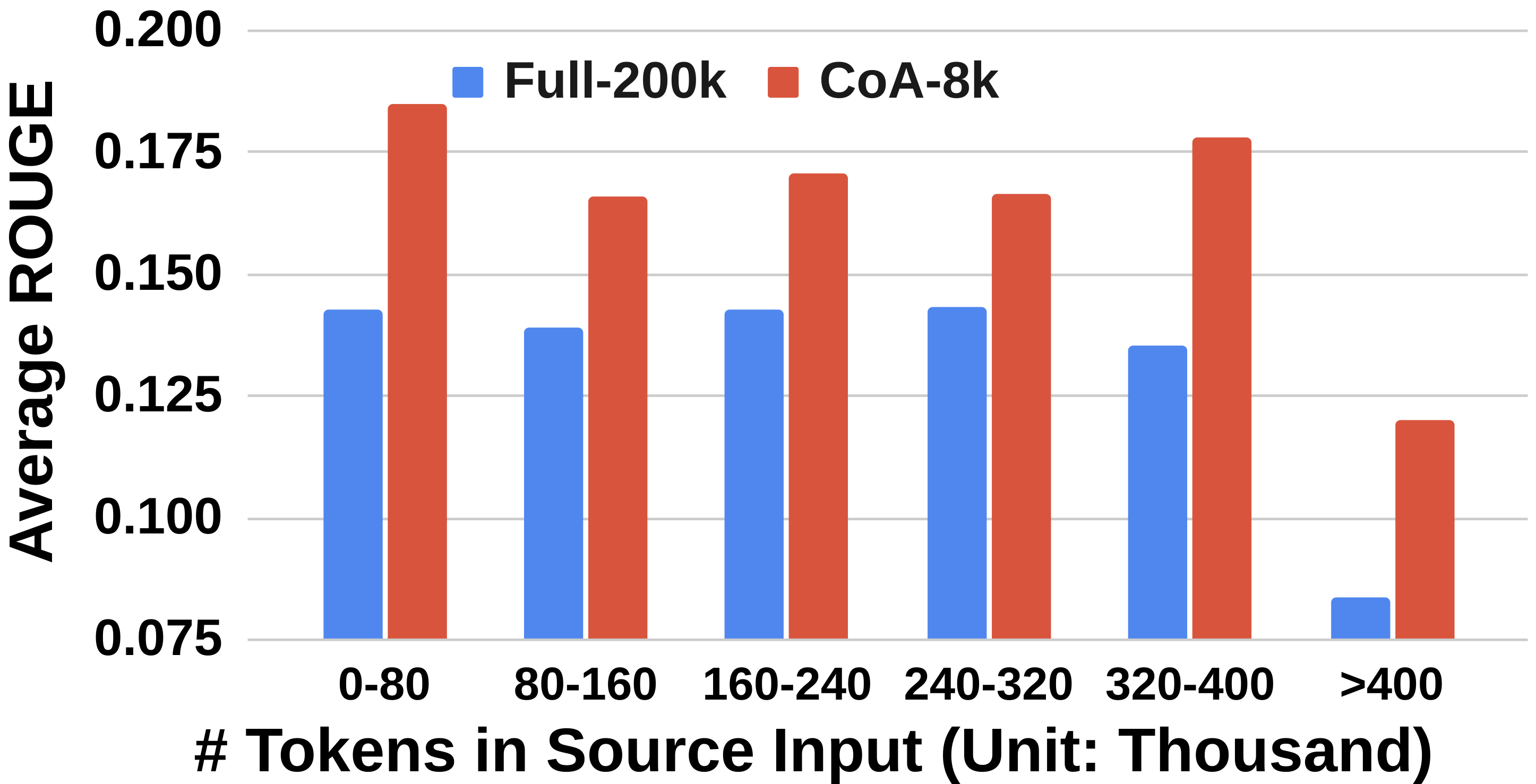}
         \caption{Claude 3 Opus}
         \label{subfig:opus}
     \end{subfigure}
        \caption{Performance of Claude 3 on BookSum. Improvement is more obvious for longer inputs.}
        \label{fig:long_context}
\end{figure}

We compare the performance of CoA and Vanilla with Claude 3 on BookSum. As shown in Figure~\ref{fig:long_context}, CoA can outperform the vanilla baseline by a large margin on various source lengths. It is worth noting that, when the length of the sample increases, the performance even increases for CoA, and the improvement over Vanilla (200k) baseline becomes more significant. The improvement of CoA reaches around 100\% when the length is larger than 400k. Thus, we can conclude that 1) CoA can still enhance the LLM performance even though the model has a very long context window limit; and 2) CoA delivers more performance gains when the input is longer.

\vspace{-2mm}
\subsection{CoA Mitigates ``Lost-in-the-Middle'' Phenomenon}
\vspace{-2mm}

To assess the ``lost-in-the-middle''~\cite{liu2024lost} effect on Vanilla and CoA models, we replicated the original study by randomly selecting 500 samples from their dataset to create a QA dataset. The results are displayed in Figure~\ref{fig:lost}. The Vanilla model exhibits a significant "lost-in-the-middle" issue, with a performance range of 6.13 ($\pm 2.17$). In contrast, CoA shows resilience against this issue, with a narrower performance gap of 4.89 ($\pm1.91$), demonstrating that CoA effectively mitigates this problem by providing each agent a shorter context to focus on.
\vspace{-2mm}
\subsection{Multi-agent Collaboration in CoA Enables Complex Reasoning over Long Context}
\vspace{-2mm}
\begin{figure}[t!]
\centering
\includegraphics[width=\linewidth]{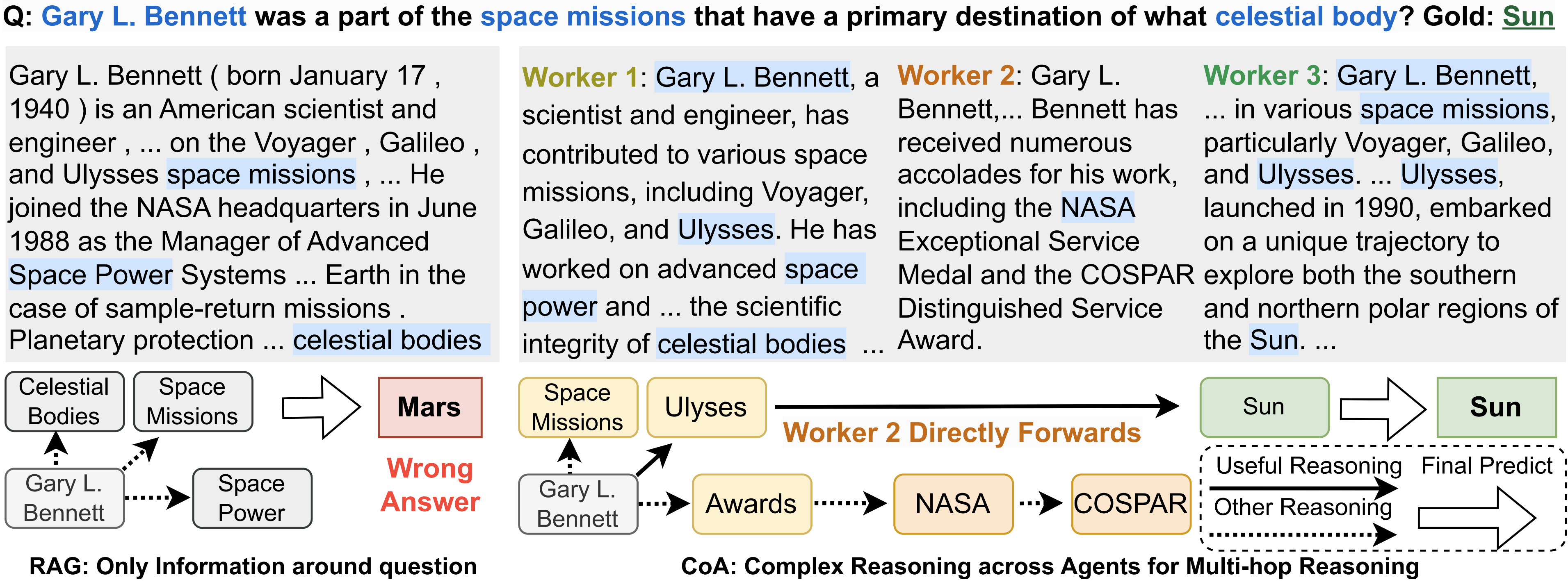}
\vspace{-3mm}
\caption{A case study of RAG (left) and CoA (right) on HotpotQA. The sequential agent communication enables CoA to perform complex multi-hop reasoning over long contexts.}
\vspace{-5mm}
\label{fig:case_study}
\end{figure}

Figure~\ref{fig:case_study} displays a sample prediction from HotpotQA. To find the correct answer, RAG retrieves text chunks with high semantic similarity with the {\it query}. However, conducting multi-hop reasoning is challenging as the critical first-hop {\it answer} often lacks semantic relevance to the {\it query}. In contrast, CoA operates differently: the first agent explores related topics without knowing the query's answer, aiding subsequent inference. The second agent, also unaware of the answer, broadens the topic scope by incorporating new information. The third agent finally discovers the answer, synthesizing information from earlier agents and new data to complete the reasoning chain. This collaborative approach highlights CoA's ability to facilitate complex reasoning across long context tasks.

\vspace{-2mm}
\subsection{Ablation Study: Effectiveness of Manager and Alternative Design Choices}
\vspace{-2mm}
\begin{table}[t!]
\caption{Ablation on CoA. Manager plays an important role in CoA, and left-to-right yields the best performance among other reading orders including Right-to-Left and Permutation.}
\resizebox{\linewidth}{!}{
\begin{tabular}{@{}lccccccc@{}}
\toprule
 & HotpotQA& MuSiQue & NarrativeQA & Qasper& QuALITY & QMSum   & RepoBench-P \\ \midrule
CoA  & 53.62 & \textbf{37.09} & \textbf{25.26} & \textbf{37.17}   & \textbf{65.42} & \textbf{16.77} & 58.25 \\
w/o Manager  & 48.58 & 26.79 & 20.80 & 29.66   & 58.80 & 16.50 & 56.16 \\
Right-to-Left    & 51.83 & 29.77 & 21.57 & 36.60   & 62.80 & 15.91 & 55.10\\
Permutation & \textbf{56.05} & 34.55 & 23.60 & 37.42   & 64.60 & 16.50 & \textbf{58.43} \\\bottomrule
\end{tabular}}
\vspace{-3mm}
\label{tab:variantion}
\end{table}

\begin{table}[t!]
\caption{Comparison of three multi-path augmentation through judge or voting. Multi-path CoA furthers enhance the performance. }
\label{tab:ensemble}
\resizebox{\linewidth}{!}{
\begin{tabular}{@{}lccccccc@{}}
\toprule
          & \multicolumn{1}{l}{HotpotQA}    & \multicolumn{1}{l}{MuSiQue}     & \multicolumn{1}{l}{NarrativeQA} & \multicolumn{1}{l}{Qasper}      & \multicolumn{1}{l}{QuALITY}     & \multicolumn{1}{l}{QMSum}       & \multicolumn{1}{l}{RepoBench-P} \\\midrule
\multicolumn{8}{l}{\textbf{Bi-direction: left-to-right and right-to-left paths (2-way)}}   \\
w/ judge & \cellcolor[HTML]{D9EAD3}54.11 & \cellcolor[HTML]{D9EAD3}36.97 & \cellcolor[HTML]{D9EAD3}24.47 & \cellcolor[HTML]{D9EAD3}39.23 & \cellcolor[HTML]{D9EAD3}65.00 & \cellcolor[HTML]{D9EAD3}16.41 & \cellcolor[HTML]{D9EAD3}49.69 \\
w/ vote  & \cellcolor[HTML]{F4CCCC}51.46 & \cellcolor[HTML]{F4CCCC}34.16 & \cellcolor[HTML]{F4CCCC}21.59 & \cellcolor[HTML]{F4CCCC}37.42 & \cellcolor[HTML]{F4CCCC}64.60 & \cellcolor[HTML]{F4CCCC}9.51  & \cellcolor[HTML]{F4CCCC}38.70 \\
oracle    & 62.12   & 48.02   & 32.84   & 46.37   & 71.80   & 18.83   & 59.67   \\ \midrule
\multicolumn{8}{l}{\textbf{Self-Consistency: five left-to-right reasoning paths (5-way)}}  \\
w/ judge & \cellcolor[HTML]{F4CCCC}57.17 & \cellcolor[HTML]{F4CCCC}38.82 & \cellcolor[HTML]{F4CCCC}21.58 & \cellcolor[HTML]{F4CCCC}36.24 & \cellcolor[HTML]{F4CCCC}62.80 & \cellcolor[HTML]{D9EAD3}17.06 & \cellcolor[HTML]{D9EAD3}46.97 \\
w/ vote  & \cellcolor[HTML]{D9EAD3}57.49 & \cellcolor[HTML]{D9EAD3}40.78 & \cellcolor[HTML]{D9EAD3}25.56 & \cellcolor[HTML]{D9EAD3}39.15 & \cellcolor[HTML]{D9EAD3}68.60 & \cellcolor[HTML]{F4CCCC}8.35  & \cellcolor[HTML]{F4CCCC}35.56 \\
oracle    & 67.07   & 55.74   & 39.89   & 52.74   & 80.40   & 20.81   & 63.52   \\\midrule
\multicolumn{8}{l}{\textbf{Permutation: five random order reasoning paths (5-way)}}          \\
w/ judge & \cellcolor[HTML]{D9EAD3}59.17 & \cellcolor[HTML]{D9EAD3}42.37 & \cellcolor[HTML]{F4CCCC}25.47 & \cellcolor[HTML]{F4CCCC}37.65 & \cellcolor[HTML]{F4CCCC}63.40 & \cellcolor[HTML]{D9EAD3}17.81 & \cellcolor[HTML]{D9EAD3}52.45 \\
w/ vote  & \cellcolor[HTML]{F4CCCC}58.29 & \cellcolor[HTML]{F4CCCC}39.17 & \cellcolor[HTML]{D9EAD3}26.58 & \cellcolor[HTML]{D9EAD3}38.09 & \cellcolor[HTML]{D9EAD3}67.60 & \cellcolor[HTML]{F4CCCC}8.31  & \cellcolor[HTML]{F4CCCC}35.44 \\
oracle    & 75.73   & 60.16   & 39.58   & 52.22   & 79.80   & 20.88   & 67.80 \\ \bottomrule
\end{tabular}}
\vspace{-3mm}
\end{table}

To demonstrate the effect of the manager, we conduct an ablation study that uses the last worker to generate results directly. As shown in Table~\ref{tab:variantion}, ``w/o Manager'' hurts the performance significantly, dropping more than 10\% on MuSiQue. This demonstrates the important role of the manager. 
Next, to empirically verify that left-to-right yields the best performance, we evaluate other orders of reading, including Right-to-Left by reading from the last chunk to the first one and Permutation which reads in random order. As shown in Table~\ref{tab:variantion}, on most of the datasets, left-to-right yields the highest score, demonstrating the advantages of natural reading order. 

\subsection{Multi-path Chain-of-Agents Further Enhances Performance}
We manually investigated the results over these three orders (left-to-right, right-to-left, permutation), and we found that other orders sometimes can produce better answers than left-to-right.
Inspired by this observation, we explore two approaches to select the best result among multiple paths. w/ vote applies majority voting over the final results while w/ judge uses an LLM to judge the most reliable $CU_l$ of diverse paths and generate the final answer. Oracle picks the best path by evaluating score of each path, yielding the upper bound performance. 
Table~\ref{tab:ensemble} compares three multi-path augmentation approaches. Surprisingly, results show that 1) all ensemble approaches (Bi-direction, Self-consistency, and Permutation) can further enhance the performance of CoA and 5-way Permutation yields the best improvement, 2) majority voting (w/ vote) of final answer is better than using an LLM as judge (w/ judge) in Self-consistency, but worse in Bi-direction, 3) using LLM judge (w/ judge) works well on long result generation tasks (QMSum, RepoBench-P), and 4) there is large space to improve because oracle (choose as answer the one with highest performance) is much higher than either w/ judge or w/ vote. We leave the direction of multi-path reasoning to future study.

\section{Conclusion}
In this paper, we propose Chain-of-Agents, a multi-agent LLM collaboration framework for solving long context tasks. It is a training free, task/length agnostic, interpretable, and cost-effective framework. Experiments show that Chain-of-Agents outperforms RAG and Long Context LLMs by a large margin despite of its simple design. Analysis shows that by integrating information aggregation and context reasoning, CoA mitigates lost-in-the-middle and performs better on longer samples. 

\paragraph{Limitations.} While CoA features with a simple and effective design, future directions can address the following limitations to further improve its prowess and efficiency. First, communication effectiveness can be further improved via finetuning or in-context learning because current LLMs are aligned with human norms which is not optimal for communication between LLMs. Second, CoA does not explore other forms of communication approaches, such as debating or complex discussions. Third, the cost and latency of running CoA can be further reduced, such as replacing some LLMs with more effective models via model routing~\cite{shnitzer2023large}.

\bibliographystyle{plain}
\bibliography{references}

%%%%%%%%%%%%%%%%%%%%%%%%%%%%%%%%%%%%%%%%%%%%%%%%%%%%%%%%%%%%

%%% END INSTRUCTIONS %%%

%%%%%%%%%%%%%%%%%%%%%%%%%%%%%%%%%%%%%%%%%%%%%%%%%%%%%%%%%%%%
\clearpage
\appendix

\section{Proof of Time Complexity}
\label{sec:proof}
Assuming the source text containing $n$ tokens, window limit of LLM is $k$ tokens, and the responses contain $r$ tokens in average. For decoder-only LLM, we grasp the operations for attention calculation as the time cost unit. Then, for Full-Context LLM, total operation for encoding input source text $T_{\text{Full}}$ is:
\begin{equation}
    T_{Enc} = (1 + 2 + \cdots + n) = \frac{(n+1)n}{2} = \mathcal{O}(n^2)
\end{equation}
Similarly, decoding starts when the model already generate all input. Thus, the first decoded token attends to $n$ positions. Total operation for decoding response is ($r\ll n$):
\begin{equation}
    T_{Dec} = (n+1 + n+2 + \cdots + n+r) = \frac{(n+1+n+r)r}{2} = \mathcal{O}(nr + r^2) = \mathcal{O}(nr)
\end{equation}

For Chain-of-Agents, we first split the source into $\lceil n/k\rceil$ chunks. Thus, total encoding time for all input is:
\begin{equation}
    T_{Enc} = (1 + 2 + \cdots + k) \times \lceil n/k\rceil = \frac{(k+1)k \times \lceil n/k\rceil }{2} = \mathcal{O}(k^2\times n/k) = \mathcal{O}(nk)
\end{equation}
Decoding starts when the model already generate $k$ tokens. Thus, the first decoded token attends to $k$ positions. Total operation for decoding response is ($r\ll k$):
\begin{equation}
    T_{Dec} = (k+1 + k+2 + \cdots + k+r)\times \lceil n/k\rceil = \frac{(k+1+k+r)r \times \lceil n/k\rceil}{2} = \mathcal{O}(nr + nr^2/k) = \mathcal{O}(nr)
\end{equation}

\section{Implementation Details}
For all experiments, we use Vertex model garden~\footnote{\url{https://cloud.google.com/model-garden}} API to use all six models. Maximum generation token is set to 2048 for gemini-ultra and set to 1024 for the rest of the models. We set temperature to 0 for all experiments except for Self-consistency setting. Table~\ref{tab:query_prompt} shows the prompt for all models. for task specific requirement of 9 datasets, we follow the original LongBench~\cite{bai2023longbench} and SCROLLS~\cite{shaham-etal-2022-scrolls}. For RAG model, we use the model provided by Huggingface\footnote{\url{https://huggingface.co/}} and run on A100 GPUs to rerank the chunks.

\begin{table*}[!t]
\centering
\small
\caption{Prompt of all models for query-based tasks.}
\begin{tabular}{lp{13cm}}
\toprule
    Vanilla & \makecell[l]{\{Task specific requirement\} \\
\{Source Input $x$ with truncation if needed\} \\
Question: \{Question $q$\} \\
Answer:}\\ \midrule
    RAG & \makecell[l]{\{Task specific requirement\} \\
\{Retrieved Chunks of Source Input $x$\} \\
Question: \{Question $q$\} \\
Answer:}\\ \midrule
    CoA & \makecell[l]{
\textbf{Worker $W_i$:} \\
\{Input Chunk $c_i$\} \\
Here is the summary of the previous source text: \{Previous Communication Unit ($CU_{i-1}$)\} \\
Question: \{Query $q$\} \\
You need to read current source text and summary of previous source text (if any) and generate a\\ summary to include them both. Later, this summary will be used for other agents to answer the\\ Query, if any. So please write the summary that can include the evidence for answering the Query: 
\\
\textbf{Manager $M$:} \\
\{Task specific requirement\} \\
The following are given passages. However, the source text is too long and has been summarized. You \\ need to answer based on the summary: \\
\{Previous Communication Unit $CU_l$\} \\
Question: \{question\} \\
Answer:
} \\

\bottomrule
\end{tabular}

\label{tab:query_prompt}
\end{table*}

\begin{table*}[!t]
\centering
\small
\caption{Prompt of all models for non-query-based tasks.}
\begin{tabular}{lp{13cm}}
\toprule
    Vanilla & \makecell[l]{\{Task specific requirement\} \\
\{Source Input $x$ with truncation if needed\} \\
Answer:}\\ \midrule
    RAG & \makecell[l]{\{Task specific requirement\} \\
\{Retrieved Chunks of Source Input $x$\} \\
Answer:}\\ \midrule
    CoA & \makecell[l]{
\textbf{Worker $W_i$:} \\
\{Input Chunk $c_i$\} \\
Here is the summary of the previous source text: \{Previous Communication Unit ($CU_{i-1}$)\} \\
You need to read the current source text and summary of previous source text (if any) and generate a \\
summary to include them both. Later, this summary will be used for other agents to generate a summary \\
for the whole text. Thus, your generated summary should be relatively long. 
\\
\textbf{Manager $M$:} \\
\{Task specific requirement\} \\
The following are given passages. However, the source text is too long and has been summarized. You \\ need to answer based on the summary: \\
\{Previous Communication Unit $CU_l$\} \\
Answer:
} \\

\bottomrule
\end{tabular}

\label{tab:nonquery_prompt}
\end{table*}

\section{Example Outputs}
\label{sec:case}
We pick one sample from HotpotQA in Table~\ref{tab:sample_hotpot}, QMSum in Table~\ref{tab:sample_qmsum}, and RepoBench-P in Table~\ref{tab:sample_repobench} to show the prediction of Chain-of-Agents, in the order of question $q$, Communication Unit $CU$ of agents, and final prediction of manager. 

\begin{table*}[!t]
\centering
\small
\caption{Example output on HotpotQA dataset.}
\begin{tabular}{lp{13cm}}
\toprule
    Question & Gary L. Bennett was a part of the space missions that have a primary destination of what celestial body? \\ \midrule
    Worker 1 &  Gary L. Bennett, a scientist and engineer, has contributed to various space missions, including Voyager, Galileo, and Ulysses. He has worked on advanced space power and propulsion systems and has been involved in planetary protection measures. Bennett's expertise has been instrumental in ensuring the scientific integrity of celestial bodies and preventing harmful contamination. He has received numerous awards and accolades for his contributions to space exploration and is recognized as a leading expert in the field of planetary protection.\\\midrule
    Worker 2 &  Gary L. Bennett, a renowned scientist and engineer, has made significant contributions to space missions, including Voyager, Galileo, and Ulysses. His expertise in advanced space power and propulsion systems, as well as planetary protection measures, has been crucial in ensuring the scientific integrity of celestial bodies and preventing harmful contamination. Bennett has received numerous accolades for his work, including the NASA Exceptional Service Medal and the COSPAR Distinguished Service Award. \\\midrule
    Worker 3 & Gary L. Bennett, a distinguished scientist and engineer, played a pivotal role in various space missions, particularly Voyager, Galileo, and Ulysses. His expertise in advanced space power and propulsion systems, coupled with his focus on planetary protection measures, has been instrumental in safeguarding the scientific integrity of celestial bodies. Ulysses, launched in 1990, embarked on a unique trajectory to explore both the southern and northern polar regions of the Sun. During its extended mission, Ulysses provided invaluable data on the Sun's magnetic field, solar wind, and the presence of dust in the Solar System. \\\midrule
    Manager & Sun \\\midrule
    RAG & Mars \\\midrule
    Gold & Sun \\
\bottomrule
\end{tabular}

\label{tab:sample_hotpot}
\end{table*}

\begin{table*}[!t]
\centering
\small
\caption{Example output on QMSum dataset.}
\label{tab:sample_qmsum}
\begin{tabular}{lp{13cm}}
\toprule
    Question & Summarize the discussion about industrial components. \\ \midrule
    Worker 1 &  The industrial designer proposed that the remote control should be made of titanium or rubber, and the buttons should be simple. The marketing expert said that the remote control should be fancy, innovative, and easy to use. They also discussed the trend of making products look like fruit and vegetables.
\\\midrule
    Worker 2 & The industrial designer proposed that the remote control should be made of titanium or 
rubber, and the buttons should be simple. The marketing expert said that the remote control should be fancy, innovative, and easy to use. They also discussed the trend of making products look like fruit and vegetables. The industrial designer said that they would explore the two options of titanium and rubber, and see if rubber is expensive. The user interface designer said that they would pretend that the modelling clay is titanium and paint it afterwards. The project manager said that they would stick with two batteries and not reinvent the wheel.
\\\midrule
    Manager & The industrial designer proposed that the remote control should be made of titanium or rubber, and the buttons should be simple. The marketing expert said that the remote control should be fancy, innovative, and easy to use. They also discussed the trend of making products look like fruit and vegetables. The industrial designer said that they would explore the two options of titanium and rubber, and see if rubber is expensive. The user interface designer said that they would pretend that the modelling clay is titanium and paint it afterwards. The project manager said that they would stick with two batteries and not reinvent the wheel. (ROUGE Score: 21.38)
 \\\midrule
    RAG &  The industrial designer and the marketing manager discussed the industrial components of the remote control. They agreed that the remote control should have a simple design, with a few buttons and a speech recognition technology. They also discussed the possibility of adding a L\_C\_D\_ display, but decided that it would be too expensive. The designer suggested that the remote control could be made of titanium or rubber, and the marketing manager agreed that a spongy design would be more appealing to consumers. (ROUGE Score: 16.54)
\\ \midrule
    Gold & The industrial designer provided several options respectively for energy, material and interface. Among these options, the industrial designer preferred traditional battery, titanium and simple push buttons so that they would have enough money for speech recognition. After that, the user interface designer proposed an LCD display only for output, which might be taken into account. The group also talked about how to enable the users to find their controllers. \\
\bottomrule
\end{tabular}
\end{table*}

\begin{table*}[!t]
\centering
\small
\caption{Example output on RepoBench-P dataset.}
\label{tab:sample_repobench}
\begin{tabular}{lp{13cm}}
\toprule
    Question & What is the next line of code: … if(type == 5 || type == 0) \{ \\ \midrule
    Worker 1 &  The method SensorDataCollectorManager.flushSensorDataCache(int type, String deviceID) flushes the database cache for the given sensor type. If the type is 0, all sensor types are flushed. The method first checks if the type is valid, and if it is, it calls the flushDBCache method for the appropriate sensor collector. SensorDataCollectorManager.flushSensorDataCache(int type, String deviceID) flushes the database cache for the given sensor type. If the type is 0, all sensor types are flushed. The method first checks if the type is valid, and if it is, it calls the flushDBCache method for the appropriate sensor collector.
\\\midrule
    Worker 2 & The method SensorDataCollectorManager.flushSensorDataCache(int type, String deviceID) flushes the database cache for the given sensor type. If the type is 0, all sensor types are flushed. The method first checks if the type is valid, and if it is, it calls the flushDBCache method for the appropriate sensor collector.
\\\midrule
    Manager & LightSensorCollector.flushDBCache(deviceID);
 \\\midrule
    RAG &  AccelerometerSensorCollector.flushDBCache(deviceID); 
\\ \midrule
    Gold & LightSensorCollector.flushDBCache(deviceID); \\
\bottomrule
\end{tabular}
\end{table*}

\section{Robustness against Context Window Size}

\begin{figure}[h!]
    \centering
    \includegraphics[width=0.5\linewidth]{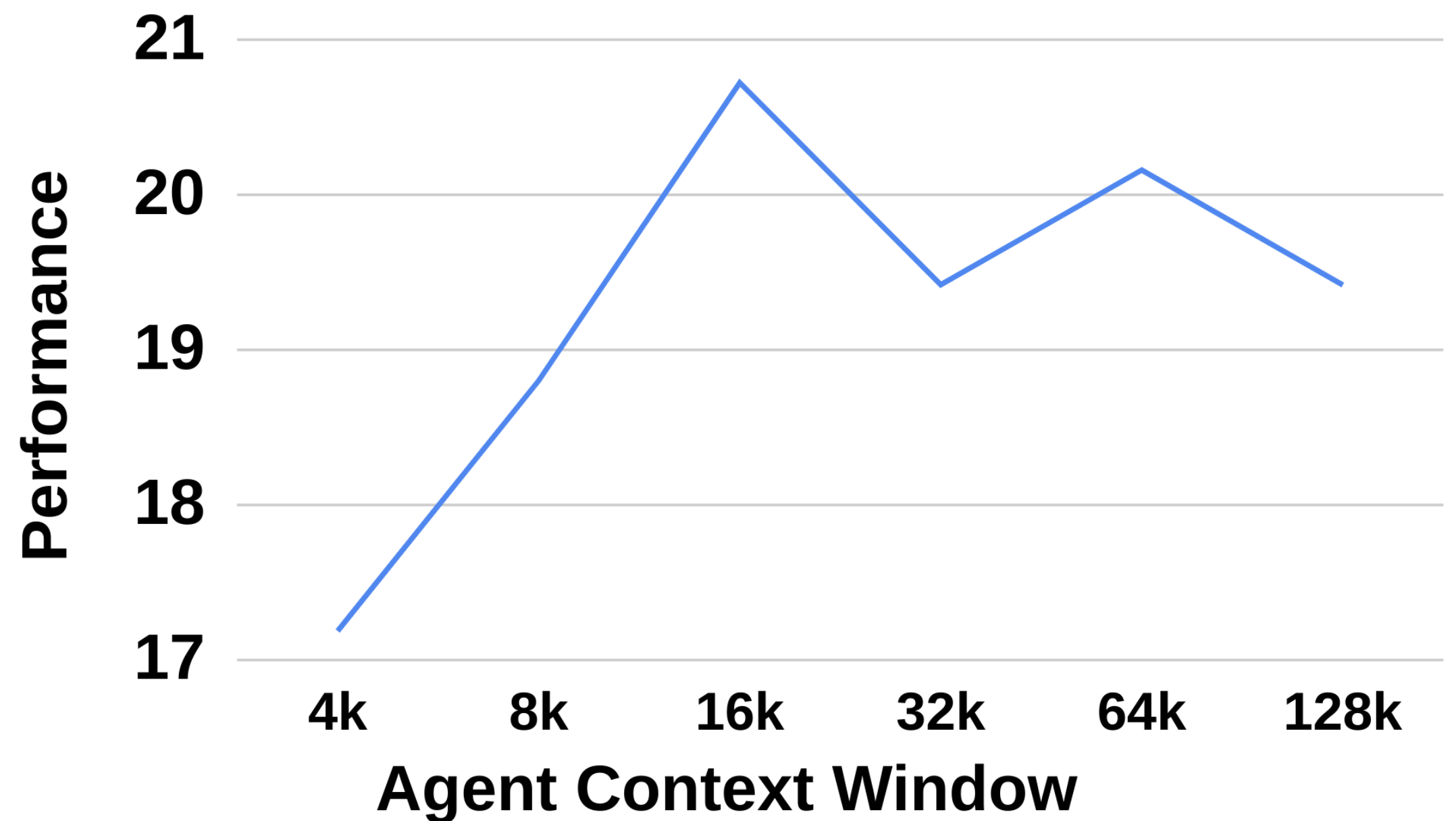}
    \caption{Performance of CoA on Claude 3 Haiku on the NarrativeQA dataset with various context window sizes of an agent. Results show the robustness of CoA towards different choices of context lengths.}
    \label{fig:context_window}
\end{figure}
We set the default context window of CoA to 8k due to the limitation of text-bison and unicorn models. To test the influence of CoA against context window change, we set window size to 4k, 8k, 16k, 32k, 64k, and 128k of Claude 3 Haiku model and evaluate on NarrativeQA dataset and see the performance change. As shown in Figure~\ref{fig:context_window}, the performance of the model increases from 4k to 16k and stabilize to around 20 with context window further increases. This result shows that CoA will  benefit from increasing length and keep stable when the length touches a bound.

\section{Broader Impacts}

Chain-of-Agents is a generic framework for long context tasks. users can apply this to diverse tasks not restricting to the mentioned ones. It will greatly increase the efficiency of individuals or companies to solve complex long context tasks. Besides, the interpretablity of such approach can reduce the misuse of the LLMs because users can check the correctness of results and decrease the possibility of making faults. However, similar to all prompt based approaches, this framework requires careful prompt design for unseen large language models, users may not get optimal solution on certain newly proposed LLMs. Besides, it may increase the number of the calls for API, causing higher network traffic and higher latency for user pools.
%%%%%%%%%%%%%%%%%%%%%%%%%%%%%%%%%%%%%%%%%%%%%%%%%%%%%%%%%%%%

\end{document}